\documentclass{article}
\usepackage{spconf,amsmath,graphicx,hyperref}
\usepackage{enumitem} 
\usepackage{booktabs} 
\usepackage{multirow}
\usepackage{graphicx}
\usepackage{tikz}
\usepackage{caption}
\usepackage{pgfplots}
\usepackage{cite}
\usepackage{tabularx}
\usepackage[table]{xcolor}

\title{DaPT: A Dual-Path Framework for \\Multilingual Multi-hop Question Answering}
\name{
\begin{tabular}{c}
Yilin Wang\textsuperscript{\rm 1},
Yuchun Fan\textsuperscript{\rm 1},
Jiaoyang Li\textsuperscript{\rm 1},
Ziming Zhu\textsuperscript{\rm 1},
Yongyu Mu\textsuperscript{\rm 1},\\
Qiaozhi He,
Tong Xiao\textsuperscript{1,2$\dagger$}, 
Jingbo Zhu\textsuperscript{1,2}
\end{tabular}
\thanks{
    $\dagger$~\ Corresponding author.} 
}
\address{\textsuperscript{1} School of Computer Science and Engineering, Northeastern University, Shenyang, China\\
\textsuperscript{2} NiuTrans Research, Shenyang, China}
\begin{document}
\ninept
\maketitle
\begin{abstract}
Retrieval-augmented generation (RAG) systems have made significant progress in solving complex multi-hop question answering (QA) tasks in the English scenario. 
However, RAG systems inevitably face the application scenario of retrieving across multilingual corpora and queries, leaving several open challenges. The first one involves the absence of benchmarks that assess RAG systems' capabilities under the multilingual multi-hop (MM-hop) QA setting. The second centers on the overreliance on LLMs’ strong semantic understanding in English, which diminishes effectiveness in multilingual scenarios.
To address these challenges, we first construct multilingual multi-hop QA benchmarks by translating English-only benchmarks into five languages, and then we propose DaPT, a novel multilingual RAG framework. DaPT generates sub-question graphs in parallel for both the source-language query and its English translation counterpart, then merges them before employing a bilingual retrieval-and-answer strategy to sequentially solve sub-questions.
Our experimental results demonstrate that advanced RAG systems suffer from a significant performance imbalance in multilingual scenarios. Furthermore, our proposed method consistently yields more accurate and concise answers compared to the baselines, significantly enhancing RAG performance on this task. For instance, on the most challenging MuSiQue benchmark, DAPT achieves a relative improvement of 18.3\% in average EM score over the strongest baseline. \footnote{The code and data will be available at https://github.com/f6ster/DaPT.}
\end{abstract}
\section{Introduction}
\label{sec:intro}
Retrieval-Augmented Generation (RAG) has emerged as a key technique for large language models (LLMs) \cite{achiam2023gpt,anthropic1claude,liu2024deepseek} to solve knowledge-intensive tasks. It introduces retrieval into the inference process, enabling models to refer to the provided relevant context and then generate the final answer, thus reducing hallucination \cite{huang2025survey, huang2025parammute, huang2025alleviating, huang2025improving}. Although the RAG paradigm performs effectively for simple, fact-based questions, it still struggles with the multi-hop question answering (QA) task. In this scenario, the clues remain in several documents, and models need to integrate the clues from the retrieved documents to form a conclusion. Related works along this line of research involve two aspects. One stream of research is structure-free optimizations of RAG systems. A prominent approach is to employ iterative retrieval to progressively refine the supporting clues found across multiple documents \cite{trivedi2022interleaving, shao2023enhancing}. Another stream of research leverages structured knowledge (e.g., trees or graphs) to either enhance the retrieval of documents dense with clues or to serve as a structural guide for the LLM to pinpoint the clues within the retrieved context \cite{edge2024local, guo2024lightrag, jimenez2024hipporag, sarthi2024raptor, gutiérrez2025ragmemorynonparametriccontinual}.

Notably, these advances have been demonstrated almost exclusively in English-centric scenarios. This presents a critical gap, as real-world applications must handle the complexities of multilingual queries and documents. This gap can be attributed to two main aspects: (i) a scarcity of multilingual multi-hop (MM-hop) benchmarks has hindered systematic evaluation, while naive 'translate-then-retrieve' strategies introduce noise and lose source-language context; and (ii) more fundamentally, the reasoning mechanisms of current methods are deeply reliant on the LLM's robust English comprehension, causing performance to degrade in cross-lingual scenarios \cite{chirkova2024retrieval, park2025investigating, qi2025consistency, 10889541, chen2024bge, fan2025slam}. 

As for aspect (i), we first constructed three new MM-hop QA benchmarks by using GPT-4o-mini to translate three mainstream QA datasets—HotpotQA \cite{yang2018hotpotqa}, 2WikiMultiHopQA \cite{ho2020constructing}, and Musique \cite{trivedi2022musique}—into five diverse languages: Swahili (Sw), Thai (Th), German (De), Spanish (Es), and Chinese (Zh). To address the issue of (ii) we seek to develop a method that can leverage the robust English comprehension of LLMs without losing crucial source-language context. To this end, we propose a \textbf{D}u\textbf{a}l-\textbf{P}a\textbf{t}h (DaPT) Framework for MM-hop QA. Our method DaPT, inspired by recent studies on enhancing model performance with cross-lingual context \cite{nie2022cross, mu2024large}, first decomposes both the source query and its English translation counterpart into parallel sub-question graphs. Then DaPT aligns and merges these graphs based on the semantic similarity of each node, yielding a unified, bilingual reasoning structure. This fused graph then enables a bilingual retrieval strategy, allowing the model to solve sub-questions sequentially by drawing evidence from a richer, cross-lingual context pool.
\begin{figure*}
\centering
\includegraphics[width=\textwidth,height=0.42\textheight,keepaspectratio]{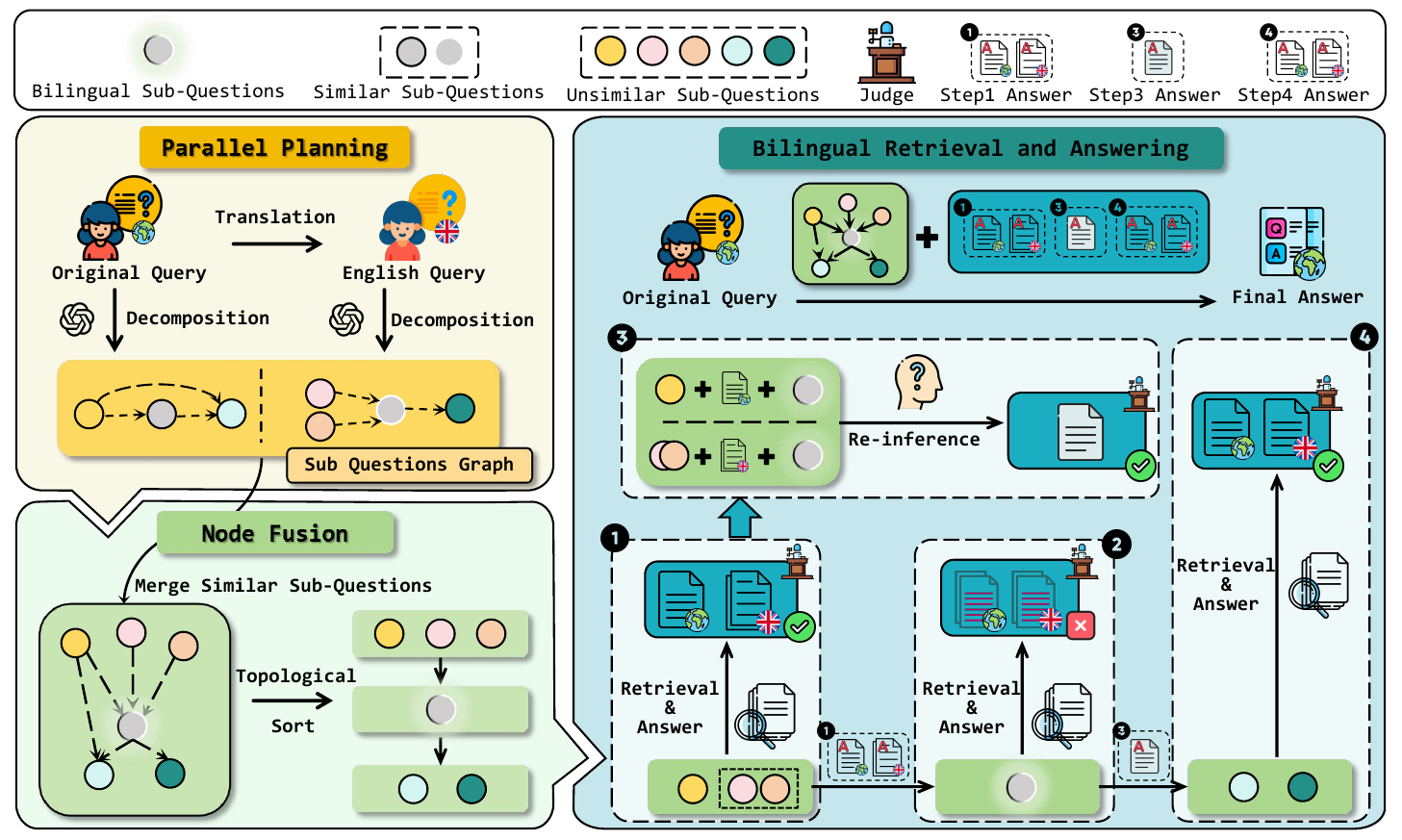}
\caption{Our proposed framework for multilingual multi-hop QA named DaPT. The framework consists of three main stages: (i) Parallel Planning: Translating the source query into English and decomposing both into two parallel sub-question graphs. (ii) Node fusion: Fusing the two graphs into a single bilingual reasoning graph based on the embedding similarity of their node of sub-questions. (iii) Bilingual Retrieval and Answering: Sequentially solving each node on the fused graph by gathering evidence with a bilingual retrieval strategy.}

\label{f1}
\end{figure*}

Extensive experimental results across three benchmarks show that DaPT achieves superior performance in multilingual multi-hop QA. For instance, DaPT raises the average EM score by 3.2 points over the HippoRAG2 on the HotpotQA benchmark. Besides, DaPT facilitates answering questions in low-resource languages, with a 4.3 F1-point improvement over HippoRAG2 in Thai on the challenging MuSiQue dataset. Moreover, DaPT demonstrates significant improvements in the Exact Match (EM) score across all benchmarks, suggesting that DaPT generates more precise and concise answers.

\section{Methodology}
\label{sec:format}
In this section, we aim to elaborate on our methodology. As shown in Fig.~\ref{f1}, we consider this a three-stage process: (1) Parallel Planning: This stage involves translating the original query into English and decomposing both versions into sub-question graphs, where each node is a single-hop question and the edges represent the dependencies between them; (2) Node Fusion: Fusing the two sub-question graphs into a single, bilingual reasoning structure by aligning nodes with high embedding similarity; and (3) Bilingual Retrieval and Answering: The nodes of the fused graph are solved sequentially. For each bilingual node, a dual-path retrieval strategy gathers context from the corpus for the LLM to generate a sub-answer. Finally, the complete reasoning path, consisting of all solved sub-questions and their answers, is synthesized to produce the final response.

\subsection{Parallel Planning}
\label{ssec:subhead}
Given an original query $q_{l}$, where $l \in \{\mathrm{sw}, \mathrm{zh}, \mathrm{de},\allowbreak \mathrm{th}, \mathrm{es}\}$, 
DaPT first translates $q_{l}$ into its English counterpart $q_{\mathrm{en}}$. 
Next, an LLM decomposes each question into an interdependent sub-question graph, yielding two separate graphs: $G_{l}$ and $G_{\mathrm{en}}$, and ensures that the resulting sub-question graphs are acyclic without circular dependencies.
Formally, each graph $G = (V, E)$ consists of node set $V = \{v_{1}, \ldots, v_{n}\}$ representing the sub-questions. 
A directed edge $(v_{i}, v_{j}) \in E$ indicates that the answer to sub-question $v_{i}$ is required to solve $v_{j}$. 
This decomposition process can be formalized as:
\begin{equation}
G_{L}(V_{L}, E_{L}) = \mathrm{Decompose}\!\left(f(q^{L})\right)
\quad L \in \{l, \mathrm{en}\},
\end{equation}
where $f(\cdot)$ is the prompting function that prepares the input question $q_{L}$ for the LLM.

As shown in Fig~\ref{f1}, the original query and its English counterpart are decomposed into two separate sub-question graphs, consisting of three and four sub-questions, respectively.
\begin{table*}[htbp]
\centering
\caption{EM and F1 scores across six languages on the three multilingual multi-hop QA datasets we constructed. Avg. denotes the average score across languages. \textbf{Bold} values indicate the best performance within each group.}
\setlength{\tabcolsep}{2.5pt}
\renewcommand{\arraystretch}{0.92}
\scriptsize
\resizebox{\textwidth}{!}{%
\definecolor{lightgray}{gray}{0.9}
\begin{tabular*}{\textwidth}{@{\extracolsep{\fill}}ll|*{7}{cc}@{}}

\toprule
\multirow{2.5}{*}{\textbf{Type}} & \multirow{2.5}{*}{\textbf{Method}} & \multicolumn{2}{c}{\textbf{Sw}} & \multicolumn{2}{c}{\textbf{Th}} & \multicolumn{2}{c}{\textbf{De}} & \multicolumn{2}{c}{\textbf{Es}} & \multicolumn{2}{c}{\textbf{Zh}} & \multicolumn{2}{c}{\textbf{En}} & \multicolumn{2}{c}{\textbf{Avg.}} \\
\cmidrule(lr){3-4} \cmidrule(lr){5-6} \cmidrule(lr){7-8} \cmidrule(lr){9-10} \cmidrule(lr){11-12} \cmidrule(lr){13-14} \cmidrule(lr){15-16}
 & & EM & F1 & EM & F1 & EM & F1 & EM & F1 & EM & F1 & EM & F1 & EM & F1 \\
\midrule

 \multicolumn{16}{c}{\textbf{HotpotQA}} \\
\midrule
\multirow{3}{*}{SF-Baselines} & Zero-shot &17.9 &28.1 &17.2 &21.8 &23.2 &32.1 &23.9 &34.3 &15.9 &33.7 &27.9 &38.4 &21.0 &31.4\\
 & CoT &16.8 &27.3 &16.7 &21.2 &24.6 &34.6 &24.3 &35.1 &15.3 &33.5 &29.8 &40.5 &21.3 &32.0 \\ & Vanilla RAG &35.1 &48.1 &38.4 &47.7 &44.0 &55.5 &42.9 &\textbf{56.2} &34.9 &35.5 &50.4 &62.3 &41.0 &50.9 \\
\midrule
\multirow{2}{*}{SA-Baselines} & GraphRAG &18.9 &24.8 &16.8 &20.6 &11.2 &14.8 &23.6 &31.3 &13.5 &23.6 &29.0 &39.3 &18.8 &25.7 \\
 & HippoRAG2 & 42.2 & \textbf{57.4} &38.9 &\textbf{49.2} &45.7 &\textbf{59.8} &46.2 &57.0 &34.6 &56.2 &54.0 &\textbf{69.0} &43.6 &\textbf{58.1} \\
 \midrule
 & DaPT & \textbf{44.5}& 53.2 &\textbf{40.7} &46.9 &\textbf{51.5} &59.5 &\textbf{50.7} &\textbf{60.3} &\textbf{35.2} &55.1 &\textbf{58.3} &67.5 &\textbf{46.8} &57.1 \\
\midrule
 \multicolumn{16}{c}{\textbf{2WikiMultiHopQA}} \\
\midrule
\multirow{3}{*}{SF-Baselines} & Zero-shot &22.5 &26.1 &27.0 &28.5 &28.4 & 32.5&27.0 &31.0 &21.5 &21.6 &26.0 &31.4 &25.4 &28.5 \\
 & CoT &16.8 &27.0 &16.8 &21.5 &27.8 &31.2 &25.7 &30.9 &23.9 &34.0 &28.8 &36.1 & 23.3 &30.1 \\ & Vanilla RAG &23.6 &37.9 &37.5 &41.0 &38.1 &43.1 &38.4 &43.5 &36.7 &36.8 &38.5 &43.7 &37.0 &41.0 \\
\midrule
\multirow{2}{*}{SA-Baselines} & GraphRAG &22.3 &25.2 &24.8 &27.1 &29.3 &32.9 &27.4 &31.4 &24.2 &32.7 &35.4 &40.1 &27.2 &31.6 \\
 & HippoRAG2 &\textbf{47.4} &\textbf{55.7} &\textbf{46.2} &\textbf{53.7} &52.7 &\textbf{61.7} &57.0 &\textbf{65.3} &30.0 &43.8 &\textbf{58.9} &\textbf{68.1} &48.7 & \textbf{58.1} \\
 \midrule
& DaPT & 46.5 & 53.1 & 45.4 & 51.6 & \textbf{55.1} & 59.1 & \textbf{57.5} & 62.8 & \textbf{39.0} & \textbf{51.1} & 56.8 & 67.2 & \textbf{50.1} & 57.5 \\
\midrule
 \multicolumn{16}{c}{\textbf{MuSiQue}} \\
\midrule
\multirow{3}{*}{SF-Baselines} & Zero-shot &\phantom{0}1.3 &\phantom{0}9.3 &\phantom{0}1.1 &\phantom{0}3.0 &\phantom{0}3.9 &11.2 &\phantom{0}2.6 &13.2 &\phantom{0}1.9 &21.1 &\phantom{0}4.9 &14.8 &\phantom{0}2.6 &12.1 \\
 & CoT &\phantom{0}1.4 &\phantom{0}9.6 &\phantom{0}1.1 &\phantom{0}3.5 &\phantom{0}3.7 &11.2 &\phantom{0}2.7 &11.9 &\phantom{0}1.7 &19.3 &\phantom{0}4.9 &14.1 &\phantom{0}2.6 &11.6 \\ & Vanilla RAG &\phantom{0}6.4 &17.2 &\phantom{0}7.1 &12.5 &13.7 &21.9 &12.5 &23.1 &\phantom{0}9.5 &28.5 &19.0 &29.1 &11.4 &22.0 \\
\midrule
\multirow{2}{*}{SA-Baselines} & GraphRAG &\phantom{0}2.3 &\phantom{0}8.2 &\phantom{0}1.7 &\phantom{0}8.1 &\phantom{0}2.2 &10.3 &\phantom{0}3.2 &13.5 &\phantom{0}2.8 &20.7 &\phantom{0}4.7 &13.3 &\phantom{0}2.8 &12.4 \\
 & HippoRAG2 &18.1 &\textbf{29.7} &12.5 &19.5 &23.8 &34.6 &22.8 &36.5 &13.3 &\textbf{33.0} &31.1 &\textbf{43.6} &20.2 &32.8 \\
 \midrule
& DaPT & \textbf{21.3} & 29.5 & \textbf{15.5} & \textbf{23.8} & \textbf{27.4} & \textbf{35.0} & \textbf{29.3} & \textbf{38.3} & \textbf{18.6} & 32.8 & \textbf{31.2} & 41.4 & \textbf{23.9} & \textbf{33.5} \\
\bottomrule
\end{tabular*}%
}
\label{t1}
\end{table*}

\subsection{Node Fusion}
\label{ssec:subhead}
Inspired by the concept of critical paths, we identify pairs of semantically similar sub-questions across the two graphs as pivotal nodes. The rationale is that while the decomposition process may vary between languages, a sub-question appearing in both graphs likely represents a robust and indispensable hub. Correctly answering these pivotal nodes is therefore critical for the success of the entire reasoning process.

In two sub-question graphs $G_{l}$ and $G_{\mathrm{en}}$, we identify the pair of nodes $(v_l^*, v_{en}^*)$ which has the highest embedding similarity across all possible pairs. This selection is formalized as:
\begin{equation}
(v_l^*, v_{en}^*) = \underset{(v_l, v_{en}) \in V^l \times V^{en}}{\arg\max} \; \text{sim}(\phi(v_l), \phi(v_{en})),
\end{equation}
where $\text{sim}(\cdot,\cdot)$ is the cosine similarity function. 

This pair is only considered for fusion if its similarity score exceeds a predefined threshold $\tau$. If no such pair exists, the process terminates.

After identifying the valid pair $(v_l^*, v_{en}^*)$, the source-language node $v_l^*$ is updated to become a bilingual node, incorporating the sub-question from its English counterpart $v_{en}^*$. All incoming and outgoing edges originally connected to $v_{en}^*$ are then redirected to the updated node $v_l^*$. Finally, the redundant English node $v_{en}^*$ is removed from the graph. This process repeats until no more pairs above the similarity threshold can be found, resulting in the final fused graph $G_{f}$.

The last step of this stage is to establish a valid execution sequence for solving the sub-questions. To achieve this, we perform a topological sort on the nodes of $G_f$. This operation yields a linear sequence of nodes $S = (v_1, v_2, \dots, v_m)$, where $m$ is the total number of nodes in the fused graph. This sequence $S$ represents the final, ordered reasoning path, which contains the updated $v_l^*$ nodes from the previous step.

\subsection{Bilingual Retrieval and Answering}
\label{ssec:subhead}
In this stage, we solve the sub-questions iteratively according to the sequence $S$. For each step, the query for the current sub-question $v_{i+1}$ is dynamically formed by combining the answer from the immediately preceding sub-question, $a_i$. This query update process is formalized as:
\begin{equation}
    q_{i+1} = \text{combine}(a_i, v_{i+1}),
\end{equation}
where $a_i$ is the answer to the previous sub-question, and $\text{combine}(\cdot)$ is a function that merges them into a new query. Subsequently, the query $q_{i+1}$ will be used to retrieve relevant context from a corpus in the same language as $q_{i+1}$, enabling the LLM to answer the sub-question $v_{i+1}$ with reference to this context and produce a response $a_{i+1}$.
 
For each bilingual node $v_l^*$, DaPT treats the bilingual sub-question as two parallel queries for retrieval, producing candidate answers from the source-language and English paths, respectively.
We employ a lightweight LLM-based verification step, denoted as $\text{Judge}(\cdot,\cdot)$, to assess whether the two candidate answers are semantically consistent and sufficient to resolve the current sub-question. If the verification succeeds, the final answer is selected via $\text{Select}(\cdot,\cdot)$, which prioritizes the source-language answer when both candidates are valid. Otherwise, DaPT triggers a self-correction mechanism $\text{ReGen}(\cdot,\cdot)$, prompting the LLM to regenerate the answer using the accumulated bilingual reasoning path as context. Formally,

\begin{equation}
a_i = 
\begin{cases}
\text{Select}(a_i^l, a_i^{\mathrm{en}}), & \text{if } \text{Judge}(a_i^l, a_i^{\mathrm{en}}) = \text{True}, \\
\text{ReGen}(v_i, \text{Path}(v_i)), & \text{otherwise},
\end{cases}
\end{equation}

This process improves robustness against retrieval noise and translation discrepancies, while maintaining a lightweight reasoning pipeline.

In Fig.~\ref{f1}, the process of solving sub-questions begins with the nodes that have no dependencies. These sub-questions serve as queries for context retrieval, after which the LLM generates the corresponding answers. Once an LLM answer is deemed correct, it becomes the answer of the previous sub-question, combining with the sub-question of step 2, serving as the query for step 2. When JudgeLLM determines that step 2 has failed, the LLM will reuse the previous reasoning path and context to attempt a new response in step 3 until it succeeds. This process will continue until all sub-questions are fully resolved. 

Finally, to alleviate the burden of long contexts and reduce potential noise, we combine all sub-questions with their corresponding answers into a reliable logical chain. This chain, together with the retrieval results of the original question, is provided to the LLM, which performs integrated reasoning to produce the final answer.

\section{Experiment}
\label{sec:pagestyle}
\subsection{Experiment Setups}
\label{ssec:subhead}
\textbf{Benchmarks}.
To evaluate the effectiveness of DaPT in MM-hop question answering, we conduct experiments on three widely used public multi-hop QA benchmarks: HotpotQA \cite{yang2018hotpotqa}, 2WikiMultiHopQA \cite{ho2020constructing}, MuSiQue \cite{trivedi2022musique}. To ensure a fair and consistent comparison across methods, we align our entire evaluation setup with that of HippoRAG\footnote{https://github.com/OSU-NLP-Group/HippoRAG}. Specifically, we adopt their methodology of using a smaller, sampled benchmark of 1,000 questions from each validation set for testing, rather than the full datasets. Furthermore, all retrieval is conducted using the exact same corpus to maintain consistency. To evaluate the performance of RAG systems in multilingual multi-hop question-answering, 
we utilized GPT-4o-mini to translate the three test datasets into five language versions. We calibrated the consistency of key entities across translations using specific rules and human verification, ensuring the integrity of multi-hop reasoning chains and the translation quality of the data.

\textbf{Baseline}.
\label{sssec:subsubhead}
Our baselines can be divided into two categories according to whether structured knowledge is retrieved in multi-hop reasoning: Structure-Free (SF) baselines and Structure-Aware (SA) baselines. The Structure-Free baselines setting includes three baselines.  Zero-shot LLM answers the question using only the model's internal knowledge, without any external corpus retrieval. Vanilla RAG is a standard RAG pipeline that uses the original question directly as a query to retrieve documents for contextual answering. CoT Prompting enhances the vanilla RAG by appending the ``Let's think step by step.'' prompt to elicit a detailed reasoning process from the model. Structure-Aware baselines setting includes two recent, high-performance graph-based RAG methods, GraphRAG \cite{edge2024local} and HippoRAG2 \cite{gutiérrez2025ragmemorynonparametriccontinual}.

\textbf{Metrics}.
By convention, we employ answer-based Exact Match (EM) score and token-based F1 score as evaluation metrics. EM score means the proportion of predictions in the test set that perfectly match the reference answer. F1 score evaluates the overlap between the predicted and reference answer tokens. We follow the implementation approach of HippoRAG, normalizing both predicted values and ground truth answers.

\textbf{Implementation details}.
\label{sssec:subsubhead}
To ensure a fair comparison, we utilize GPT-4o-mini as the default LLM across all methods. Based on previous work\cite{chirkova2024retrieval, park2025investigating}, we use the advanced and available BGE-m3\footnote{https://huggingface.co/BAAI/bge-m3} as the embedding model across all methods. For the Top-k parameter of each method, we set $k=5$ for HippoRAG2 and $k=3$ for all other baselines. The fusion threshold $\tau$ is set to 0.8.  All experiments are conducted on 8*NVIDIA GeForce RTX 3090 GPUs.

\subsection{Main Results}
\label{ssec:subhead}
\textbf{Advanced RAG systems exhibit significant performance imbalances in multilingual scenarios.}  As shown in Table~\ref{t1}, while the HippoRAG2 achieves a strong performance of 54.0 EM score on the HotpotQA and 31.1 on the harder benchmark MuSiQue in English, its performance significantly declines in moderately low-resource languages, such as Thai. Specifically, it shows a decrease of 28.0\% (54.0 $\rightarrow$ 38.9) on the HotpotQA and 59.8\% (31.1 $\rightarrow$ 12.5) on MuSiQue test sets.

\textbf{DaPT enhances multilingual performance on MM-hop question answering.} As shown in Table~\ref{t1}, DaPT demonstrates a consistent advantage over all baselines on the EM score across the three MM-hop datasets. In particular, DaPT achieves average improvements of 6.8\% on HotpotQA and 15.5\% on MuSiQue compared with HippoRAG2. These results highlight the effectiveness of DaPT in solving MM-hop question answering. Remarkably, DaPT leads to substantial gains for high-resource languages, with improvements of 4.6\% (52.7$\rightarrow$ 55.1) in German and 30\% (30.0$\rightarrow$ 39.0) in Chinese compared with HippoRAG2. This suggests that DaPT, by decomposing the query into a sub-question graph and fusing it with their parallel English versions for reasoning, effectively bridges the language gap to some extent. In addition, we note that  DaPT does not always achieve the highest F1 score, which suggests its preference for offering the most concise and precise responses.

\begin{table}[htbp]
\centering
\caption{Ablation performance on multilingual HotpotQA. EM and F1 represent the average scores for respective non-English scores.}
\label{tab:t2}
\begin{tabular}{lcc}

\toprule
\textbf{Method} & \textbf{EM} & \textbf{F1} \\
\midrule
DaPT (Full Model) & \textbf{44.5} & \textbf{55.1} \\
\midrule
w/o Decomposition & 28.3 (↓16.2) & 46.8 (↓8.2) \\
w/o Fusion & 42.7 (↓1.9) & 53.1 (↓1.9) \\
\midrule
Translate Q\&A & 41.4 ({↓3.2}) & 51.5 (↓3.5) \\
\bottomrule
\end{tabular}

\end{table}
\subsection{Ablation Study}
\label{ssec:subhead}
We present extensive ablation studies of DaPT on the effectiveness of decomposition and fusion frameworks. Our ablation study on the multilingual HotpotQA dataset investigates the impact of DaPT's decomposition and fusion framework. We compare DaPT with three variants: (i) DaPT removes the decomposition stage, (ii) DaPT removes the graph fusion stage (iii) Translate Q\&A. The Translate Q\&A means a simple method where the source query is translated to English for the entire reasoning process, and the final answer is translated back to original language. The performance of these ablated methods on English queries would be redundant: the variant without decomposition behaves identically to the Vanilla RAG baseline in English, while the other two variants are identical to the full DaPT framework when handling English questions. We only report the average EM scores and F1 scores on the non-English languages.

As shown in Table~\ref{tab:t2}, DaPT achieves the best performance compared to the framework without its decomposition and fusion components and the Translate Q\&A method. The results demonstrate the effectiveness and the necessity of the decomposition and fusion framework within DaPT. Furthermore, it confirms that our method is inherently more advanced than a simple Translate Q\&A strategy.

\section{Conclusion}
\label{sec:typestyle}
In this work, we address the challenge of multilingual multi-hop QA with two key contributions. First, we construct a new set of multilingual multi-hop benchmarks by translating three mainstream benchmarks, HotpotQA, 2WikiMultiHopQA, MuSiQue into five diverse languages, ranging from low- to high-resource. Second, we propose DaPT, a novel multilingual RAG framework for MM-hop question answering. DaPT first translates the original-language query into a parallel English counterpart, then decomposes and fuses both into a bilingual sub-question graph, and finally performs sequential reasoning over this unified structure leading to richer reasoning paths, more accurate results, and better interpretability. Extensive experiments on our benchmarks demonstrate that DaPT consistently outperforms structure-aware approaches. 

\section{Acknowledgements}
This work was supported in part by the National Science Foundation of China (Nos. 62276056 and U24A20334), the Yunnan Fundamental Research Projects (No.202401BC070021), the Yunnan Science and Technology Major Project (No. 202502AD080014), the Fundamental Research Funds for the Central Universities (Nos. N25BSS054 and N25BSS094), and the Program of Introducing Talents of Discipline to Universities, Plan 111 (No.B16009).




\bibliographystyle{IEEEbib}
\bibliography{refs}

\end{document}